# ONTOLOGY-BASED KNOWLEDGE REPRESENTATION FOR BONE DISEASE DIAGNOSIS: A FOUNDATION FOR SAFE AND SUSTAINABLE MEDICAL ARTIFICIAL INTELLIGENCE SYSTEMS

Loan Dao[1], Ngoc Quoc Ly[2]

[1, 2] *Dept. of Computer Vision and Cognitive Cybernetics*

*University of Science, VNUHCM, Ho Chi Minh, Vietnam*

*Viet Nam National University, Ho Chi Minh City, Vietnam*

Email: [1] 24N11102@student.hcmus.edu.vn, [2] lqngoc@fit.hcmus.edu.vn

**ABSTRACT**

Medical artificial intelligence (AI) systems frequently lack systematic domain expertise integration, potentially compromising diagnostic reliability. This study presents an ontology-based framework for bone disease diagnosis, developed in collaboration with Ho Chi Minh City Hospital for Traumatology and Orthopedics. The framework introduces three theoretical contributions: (1) a hierarchical neural network architecture guided by bone disease ontology for segmentation-classification tasks, incorporating Visual Language Models (VLMs) through prompts, (2) an ontology-enhanced Visual Question Answering (VQA) system for clinical reasoning, and (3) a multimodal deep learning model that integrates imaging, clinical, and laboratory data through ontological relationships. The methodology maintains clinical interpretability through systematic knowledge digitization, standardized medical terminology mapping, and modular architecture design. The framework demonstrates potential for extension beyond bone diseases through its standardized structure and reusable components. While theoretical foundations are established, experimental validation remains pending due to current dataset and computational resource limitations. Future work will focus on expanding the clinical dataset and conducting comprehensive system validation.

## 1. INTRODUCTION

Bone pathology diagnosis in medical imaging faces significant challenges in integrating structured clinical knowledge with artificial intelligence capabilities. Recent progress in medical image segmentation (Dao & Ly, 2023) and classification (Dao & Ly, 2024) highlights a critical gap in knowledge integration across various diagnostic tasks. Traditional AI approaches often operate as "black boxes," lacking transparent integration of medical expertise and standardized clinical knowledge. This limitation becomes particularly critical in bone disease diagnosis, where complex relationships between anatomical structures, pathological patterns, and clinical presentations require systematic domain knowledge incorporation.

The integration of VLMs has transformed traditional single-modal approaches in medical imaging analysis (Mohammed et al., 2023). However, current VLM applications in bone disease diagnosis lack structured knowledge representation, potentially leading to inconsistent or unreliable diagnostic suggestions. Recent studies by Ghidalia et al. (2024) emphasize the importance of combining ontology with machine learning for encoding complex medical relationships in machine-readable formats. This research addresses these challenges through a novel bone disease ontology framework that systematically integrates clinical expertise with AI capabilities.

The framework's significance lies in three key theoretical contributions. First, it introduces an ontology-guided hierarchical neural network that combines VLM capabilities with traditional computer vision tasks, building upon recent advances in automated ontology construction (Zengeya & Fonou-Dombeu,

2024). Second, it develops an ontology-enhanced VQA system that maintains clinical interpretability through standardized medical terminology, addressing challenges identified by Remy et al. (2024) in medical knowledge representation. Third, it presents a multimodal deep learning architecture that integrates imaging, clinical, and laboratory data through ontological relationships, extending the multimodal integration framework proposed by Franciosi et al. (2024).

This research establishes a theoretical foundation for knowledge-driven medical AI systems. The framework's modular design and standardized structure support extension beyond bone diseases, offering potential applications in various medical domains. The methodology emphasizes clinical safety and reliability through systematic knowledge digitization and standardized terminology mapping, addressing concerns raised by Jing et al. (2023) regarding Clinical Decision Support Systems adaptability.

The subsequent sections detail the theoretical framework, methodology, and implementation approach. The literature review examines current advances in medical ontologies and AI integration, particularly focusing on rare disease diagnosis approaches highlighted by Roman-Naranjo et al. (2023). The methods section presents the ontology development process and its application in three key diagnostic tasks. Due to current dataset and computational resource limitations, experimental validation remains as future work, focusing on expanding the clinical dataset and conducting comprehensive system testing.

## 2. LITERATURE REVIEW, THEORETICAL FRAMEWORK AND METHODS

### 2.1. LITERATURE REVIEW

The integration of ontology and deep learning is transforming medical diagnosis systems. Recent advancements highlight deep learning's ability to process diverse data types, from medical imaging to molecular biology, in omics data analysis and clinical diagnostics (Mohammed et al., 2023). Ontologies are key in organizing medical knowledge, offering structured frameworks for clinical decision support. Combining ontology with machine learning has led to new paradigms in building decision support systems, especially for encoding complex medical relationships in machine-readable formats (Ghidalia et al., 2024).

The combination of ontology and deep learning is especially valuable in rare disease diagnosis. A systematic review by Roman-Naranjo et al. (2023) highlights the importance of integrating ontologies like the Human Phenotype Ontology with AI algorithms to improve diagnostic accuracy in rare genetic diseases. Their findings show notable improvements in identifying complex phenotype patterns. In Clinical Decision Support Systems (CDSS), machine learning combined with ontological frameworks enhances adaptation to complex clinical scenarios, resulting in more accurate and reliable diagnoses (Jing et al., 2023).

Recent advances in automated ontology construction have broadened healthcare applications. Zengeya and Fonou-Dombeu (2024) review deep learning methods for building complex ontologies from raw data. The trend toward multimodal integration is highlighted by Franciosi et al. (2024), who propose a framework combining multiple data sources for diagnosis and health monitoring.

Challenges and Research Gaps

- Ontology Digitization Challenges: Medical ontology digitization faces significant challenges due to rapid developments in medical knowledge representation (Remy et al., 2024). The integration of clinical knowledge graphs with large language models presents particular challenges in bone pathology representation. When integrating SNOMED-CT ontology with massive language models, Remy et al. (2024) found that maintaining medical accuracy required specialized methodologies. The continuous expansion of medical knowledge necessitates regular updates to ontological frameworks, and integrating clinical data from multiple sources remains challenging.

- VLMs in Medical Imaging: Hu et al. (2024) explored the limitations of VLMs in medical imaging. Current models struggle with domain-specific context and require specialized clinical training. Additionally, improving language-based interpretation accuracy remains essential for clinical VLM implementation.
- Multimodal Learning Barriers: Mou et al. (2024) presented a framework for integrating multiple modalities in medical image processing, addressing challenges in synchronization and data consistency. Their research indicates that missing or inconsistent data significantly affects multimodal model performance, highlighting the need for robust feature alignment.
- Clinical Implementation Gaps: Huemann et al. (2024) investigated the gaps in using VLMs in healthcare. They identified issues with real-time performance, workflow integration, and regulatory compliance. The authors emphasized the importance of understanding model operations and implementing effective validation techniques.

Research Directions

- Recent investigations suggest promising development opportunities:
- Self-supervised learning can enhance healthcare VLMs (Hu et al., 2024)
- Dynamic ontology updates using knowledge graphs (Remy et al., 2024)
- Federated learning for multi-center collaboration
- Explainable AI for improved clinical decision-making

Literature Review Summary: Current research highlights significant limitations in bone pathology AI systems. Hu et al. (2024) emphasize the need for medical expertise integration in deep learning models, while Mou et al. (2024) address challenges in processing multimodal medical data. Huemann et al. (2024) discuss deployment and validation issues. Together, these findings advocate for an ontology-driven approach to integrating multimodal data in clinical setting.

## 2.2. THEORETICAL FRAMEWORK

The theoretical foundation of this research combines ontological knowledge representation, multimodal learning architectures, and VLMs, forming a comprehensive framework for medical diagnosis systems.

### 2.2.1. Knowledge Representation and Ontological Modeling

The framework uses Description Logic (DL), specifically the ALC variant, as its formal foundation. The ontological structure $O = (C, R, A)$ includes concept sets $C$, relation sets $R$, and axiom sets $A$, enabling the representation of medical knowledge through:

- Hierarchical Disease Classification and Malignancy:

$$D_i \sqsubseteq D_j \sqcap M \qquad (1)$$

  where:
  $D_i$ represents a specific disease category
  $D_j$ represents its parent category
  M represents malignancy status (benign/malignant)
  This formalization ensures automatic malignancy determination upon disease classification (Zhang et al., 2024).

- Multimodal Feature Integration:

$$F \sqcap \exists hasFeature.(I \sqcup L \sqcup T) \qquad (2)$$

  where: $F$: Feature space; $I$: Image modalities; $L$: Laboratory results; $T$: Text descriptions
  This enables integration of visual, numerical and textual data (Liu et al., 2024).

### 2.2.2. Multimodal Neural Network Architecture

The neural network architecture uses a multimodal approach through:

- Vision-Language Integration:

$$VLM(x, p) = Transformer(f_v(x), f_t(p)) \in \mathcal{R}^d \qquad (3)$$

  where:

- $f_v$: CNN encoder mapping images to $\mathcal{R}^m$
- $f_t$: BERT encoder mapping text to $\mathcal{R}^n$
- Transformer: mapping from ($\mathcal{R}^m$, $\mathcal{R}^n$) to $\mathcal{R}^d$
- $d$: dimension of the combined feature space

- Ontology-Guided Attention:

$$Attention(Q, K, V) = softmax(\frac{QK^T}{\sqrt{d_k}})V \tag{4}$$

where:

- $Q \in \mathcal{R}^{n \times d_k}$ (Query matrix)
- $K \in \mathcal{R}^{m \times d_k}$ (Key matrix)
- $V \in \mathcal{R}^{m \times d_v}$ (Value matrix)
- $d_k$: dimension of query/key vectors
- softmax: applied row-wise
- Output $\in \mathcal{R}^{n \times d_v}$

This mechanism aligns learned features with domain knowledge.

### 2.2.3. Task-specific components

The multimodal integration framework implements specialized components for different diagnostic tasks. These components leverage the ontological structure and VLM capabilities to process various medical data types effectively. Table 1 presents the three primary diagnostic tasks, illustrating their distinct VLM roles, ontological functions, and processing mechanisms within the integrated framework.

*Table 1. Diagnostic Tasks and Their Characteristics in the Integrated Framework*

| Diagnostic Task | VLM Role | Ontology Function | Processing Mechanism |
|---|---|---|---|
| Segmentation-Classification | Image-prompt pair processing | Anatomical region identification | Disease and malignancy determination |
| Visual Question Answering | Natural language interaction | Medical knowledge structuring | Multi-head attention reasoning |
| Multimodal Deep Learning | Cross-modal feature extraction | Feature constraint and knowledge fusion | Hierarchical classification with parallel data processing |

Key Framework Capabilities:
- Automated ontology digitization through VLM integration
- Cross-modal feature learning and alignment
- Knowledge-guided diagnostic reasoning

Note: The framework's experimental validation remains as future work due to current data collection and computational resource constraints. Future research will focus on expanding the clinical dataset and conducting comprehensive testing to validate the theoretical foundations presented.

### 2.3. METHODS

### 2.3.1 Ontology Development Framework

A robust ontology development framework is essential for organizing medical knowledge and building effective healthcare information systems. Fig. 1 illustrates a comprehensive architectural framework with three primary components:

*A. Core Architecture* - A three-layer hierarchical structure that forms the foundation, progressing from core concepts to specific applications.

*B. Knowledge Representation Structure* - A detailed organization of medical knowledge, divided into core and supporting components, each targeting specific aspects of medical information management.

*C. Implementation for Bone Diseases* - A practical example demonstrating the framework's implementation for bone diseases, covering disease classification, anatomical mapping, and clinical correlations.

The framework emphasizes theoretical rigor and practical applicability, facilitating the integration of complex medical concepts while allowing for domain-specific adaptations. This approach ensures scalability, interoperability, and sustainable evolution of healthcare knowledge management systems.

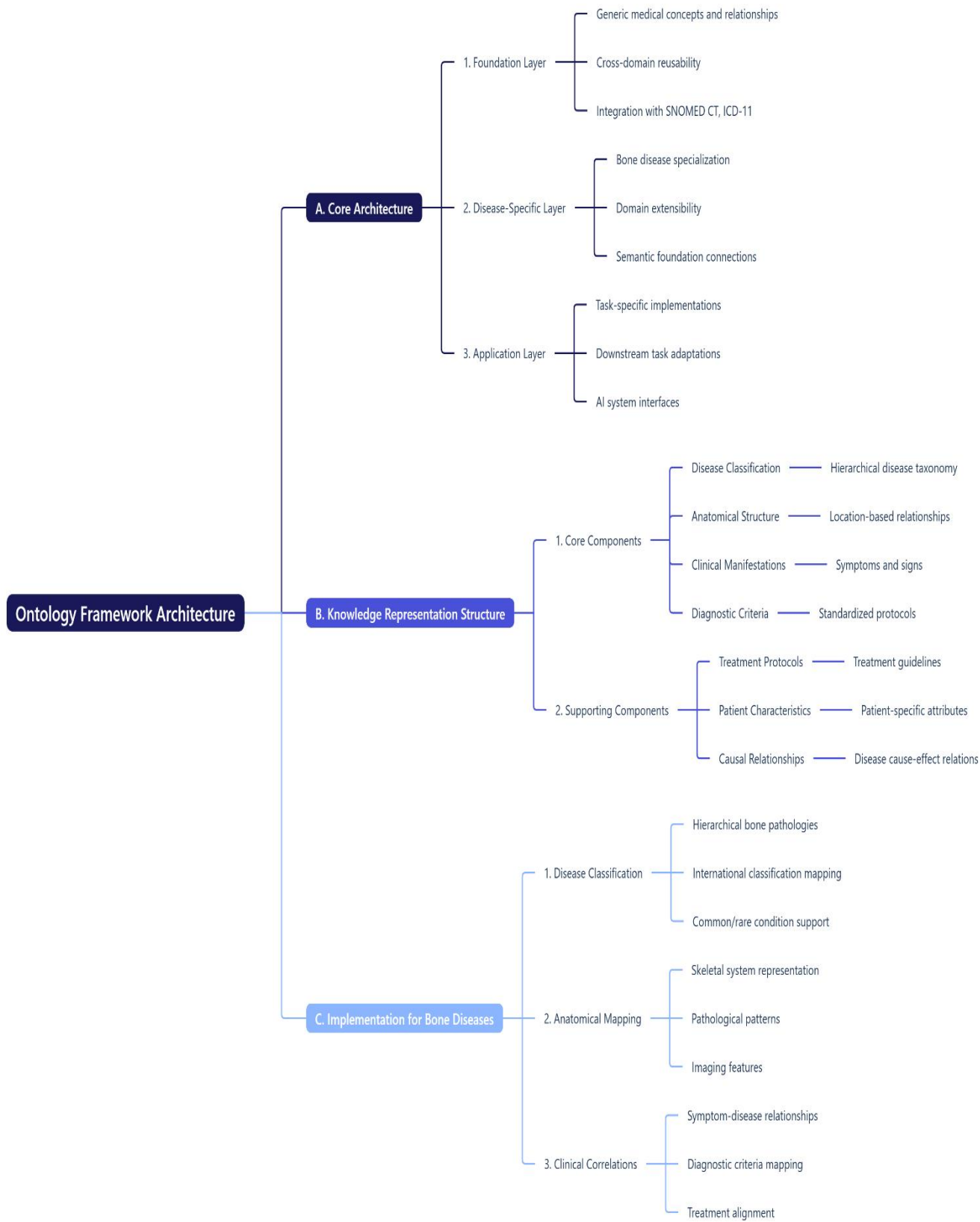

Fig. 1. Ontology Framework Architecture for Medical Knowledge Organization

## 2.3.2 Ontology Development Process

The framework's core structure consists of seven primary classes (Disease, AnatomicalStructure, Diagnosis, Symptom, Treatment, Patient, and Cause) with their corresponding subclasses (Table 2). These classes are interconnected through object properties (hasLocation, hasDiagnosis) and data properties (hasName, hasDescription), with key relationships detailed in Table 3.

*Table 2. Primary Classes of the Bone Disease Ontology and Their Significance*

| Class Name | Significance |
|---|---|
| Disease | It supports accurate classification and differential diagnosis in AI applications through a hierarchical structure, covering a wide range of pathologies, including BoneDisease, from common to rare genetic disorders. |
| AnatomicalStructure | It enables accurate localization of pathologies and understanding of disease progression through a detailed representation of the skeletal system, including bones, joints, and soft tissue. |
| Diagnosis | It enhances interoperability between clinical systems and AI models by standardizing diagnostic information, procedures, criteria, and classification systems for bone disease identification |
| Symptom | It supports pattern recognition in AI-driven diagnostic systems by offering a comprehensive catalog of bone disease manifestations, including both common and rare presentations. |
| Treatment | It enables AI systems for personalized treatment planning by structuring information on therapeutic approaches, including pharmacological, surgical, and rehabilitative interventions. |
| Patient | It allows AI models to account for individual variability by integrating patient-specific information, such as demographics, medical history, and risk factors. |
| Cause | It aids in understanding disease etiology and risk assessment by documenting genetic, environmental, and lifestyle factors contributing to bone diseases. |

*Table 3. Key Properties and Relationships in the Bone Disease Ontology*

| Property Type | Property Name | Domain | Range | Significance |
|---|---|---|---|---|
| Object Property | hasLocation | BoneDisease | AnatomicalStructure | It enables precise localization of pathologies within the skeletal system. |
| | hasDiagnosis | BoneDisease | Diagnosis | It links diseases to standardized diagnostic criteria, ensuring consistent classification. |
| | hasSymptom | BoneDisease | Symptom | It facilitates symptom-based reasoning and pattern recognition in AI models. |
| | hasTreatment | BoneDisease | Treatment | It Supports the development of treatment recommendation systems. |
| | isPartOf | AnatomicalStructure | AnatomicalStructure | It represents hierarchical relationships within the skeletal system, essential for understanding disease spread and impact. |
| Data Property | hasName | Thing | string | It ensures consistent naming across the ontology, supporting natural language processing applications. |
| | hasDescription | Thing | string | It provides detailed, human-readable information for each entity, enhancing interpretability. |
| | hasOnsetDate | BoneDisease | dateTime | It enables temporal analysis of disease progression and treatment efficacy. |
| | hasSeverity | Symptom | {mild, moderate, severe} | It supports quantitative assessment of disease impact and treatment urgency. |
| | hasFrequency | Symptom | {rare, occasional, frequent, constant} | It aids in characterizing symptom patterns, essential for accurate diagnosis. |

Clinical reasoning is implemented via SWRL rules, exemplified by automated fracture diagnosis:

BoneDisease(?d) ^ hasSymptom(?d, Pain) ^ hasLocation(?d, FemoralNeck) ^ hasAgeGroup(Patient, Elderly) -> suspectDiagnosis(?d, FemoralNeckFracture)

Additional rules and their clinical applications are outlined in **Error! Not a valid bookmark self-reference.**.

The ontology's reasoning foundation is built on three axiom types (detailed in Table 5): class axioms for hierarchies, property axioms for relationships, and individual axioms for clinical protocols, enabling robust automated diagnostic support in orthopedics.

*Table 4. SWRL Rules and Their Applications in AI-Based Bone Disease Management*

| Rule Category | SWRL Expression | AI Application Domain | Clinical Purpose | Implementation Significance |
|---|---|---|---|---|
| Analgesic Treatment Rule | Patient(?p) ^ hasSymptom(?p, ?pain) ^ Pain(?pain) ^ hasSymptom(?p, ?swelling) ^ Swelling(?swelling) ^ Analgesics(?analgesics) -> requiresTreatment(?p, ?analgesics) | Classification Systems | Pain Management | Automatically identifies cases needing analgesic intervention based on symptom patterns and severity |
| Osteoarthritis Diagnosis Rule | BoneDisease and (hasSymptom some Pain) and (hasSymptom some LimitedMobility) and (hasAge some xsd:integer[>= 50]) | Multimodal Deep Learning | Disease Diagnosis | Combines age, mobility limitations, and pain patterns for accurate osteoarthritis detection |
| Fracture Treatment Rule | Fracture(?f) ^ hasLocation(?f, ?loc) ^ hasSymptom(?f, ?pain) ^ Pain(?pain) ^ Immobilization(?immob) -> requiresTreatment(?f, ?immob) | Image Segmentation | Fracture Management | Links anatomical locations to corresponding immobilization protocols systematically. |
| Post-Surgery Treatment Rule | Surgery(?s) ^ treatedBy(?p, ?s) ^ PhysicalTherapy(?pt) -> requiresTreatment(?p, ?pt) | Visual Question Answering | Rehabilitation Planning | Standardizes rehabilitation protocols based on surgical types. |
| Osteomyelitis Treatment Rule | Osteomyelitis(?o) ^ Antibiotics(?a) -> requiresTreatment(?o, ?a) | Classification Systems | Infection Management | Standardizes antibiotic protocols for bone infections based on pathogen patterns. |

*Table 5. Axiom examples*

| Axiom Type | Example | Meaning |
|---|---|---|
| Class Axiom | Osteosarcoma ⊑ MalignantTumor | Osteosarcoma is a type of malignant tumor |
| Property Axiom | hasLocation ∘ isPartOf ⊑ hasLocation | Location transitivity |
| Individual Axiom | Fracture(x) ∧ hasLocation(x,Femur) → requiresImaging(x,XRay) | Fractures in femur require X-ray |

Additional Notes:

- All properties support inheritance through the class hierarchy
- Properties can have multiple domains and ranges
- Many properties have inverse relationships defined
- Temporal aspects are captured through data properties
- The ontology supports both qualitative and quantitative data

### 2.3.3. Practical Applications of the Ontology Framework

This section outlines the implementation of the Bone Disease Ontology in key medical AI applications, validated by clinical cases from Ho Chi Minh City Hospital for Traumatology and Orthopedics.

A. *Technical Environment*
- Programming Language: Python 3.8
- Ontology Development: Protégé 5.6.4
- Key Libraries:
    - pandas 1.4.2 (data processing)
    - scikit-learn 0.24.2 (machine learning)
    - NLTK 3.6.5 (text processing)
    - Owlready2 0.36 (ontology manipulation)

B. *Data Collection and Distribution*

Dataset characteristics from Ho Chi Minh City Hospital of Traumatology and Orthopedics (2023):
- Total cases: 1,247
- Data format: Electronic health records and images
- Disease distribution: Trauma: 876 cases (70.2%); Tumor: 241 cases (19.3%); Inflammatory: 130 cases (10.5%)
  
  The distribution reveals that trauma cases dominate the dataset (70.2%), which aligns with the hospital's primary focus on traumatology. This high percentage of trauma cases suggests the need for specialized trauma care protocols and resources in urban healthcare settings.
- Anatomical Distribution: Lower Limb: 658 cases (52.8%); Upper Limb: 389 cases (31.2%); Trunk: 200 cases (16.0%)
  
  The anatomical distribution shows a predominance of lower limb cases (52.8%), particularly affecting the femur and tibia. This pattern may be attributed to the higher vulnerability of lower extremities in traffic accidents and falls, which are common in urban environments.

Note: The current dataset serves as a preliminary foundation. Future work will focus on expanding the dataset size and diversity to enable comprehensive system validation and testing.

C. *Text Processing Algorithm*

Pseudocode for diagnosis standardization:

```
Algorithm 1: Medical Text Normalization
Input: Raw diagnostic text
Output: Standardized medical terms
Steps:
1. Remove special characters
2. Convert to lowercase
3. Tokenize text
4. Map to standard terminology
```

The methodology prioritizes a comprehensive approach to medical knowledge management through systematic organization and standardized terminology mapping. The implementation architecture emphasizes modularity and scalability, ensuring adaptability to various clinical scenarios and healthcare environments.

This research framework sets the foundation for future developments, which will focus on expanding the clinical dataset and conducting thorough system validation. The implementation roadmap includes comprehensive testing protocols, clinical validation procedures, and performance optimization strategies to ensure robust and reliable system operation in real-world medical settings.

### 2.3.4. Ontology Applications in Downstream Tasks

A visual example of applying ontology in the problem of classifying bone diseases. The BoneDx Ontology (Fig. 2) structures bone pathology diagnosis through three interconnected branches: Symptoms, Causes, and AnatomicalStructure. Consider a practical case from the Bone Disease dataset: a patient presents with "Pain" in "LowerLimb", lifestyle factors indicate "PhysicalActivity" strain, and "XRay" examination reveals "CartilaginousTumor" classified as "BenignTumor". This ontology-guided pathway demonstrates how clinical data is systematically processed from symptoms to diagnosis, while enforcing standard medical protocols in AI decision-making.

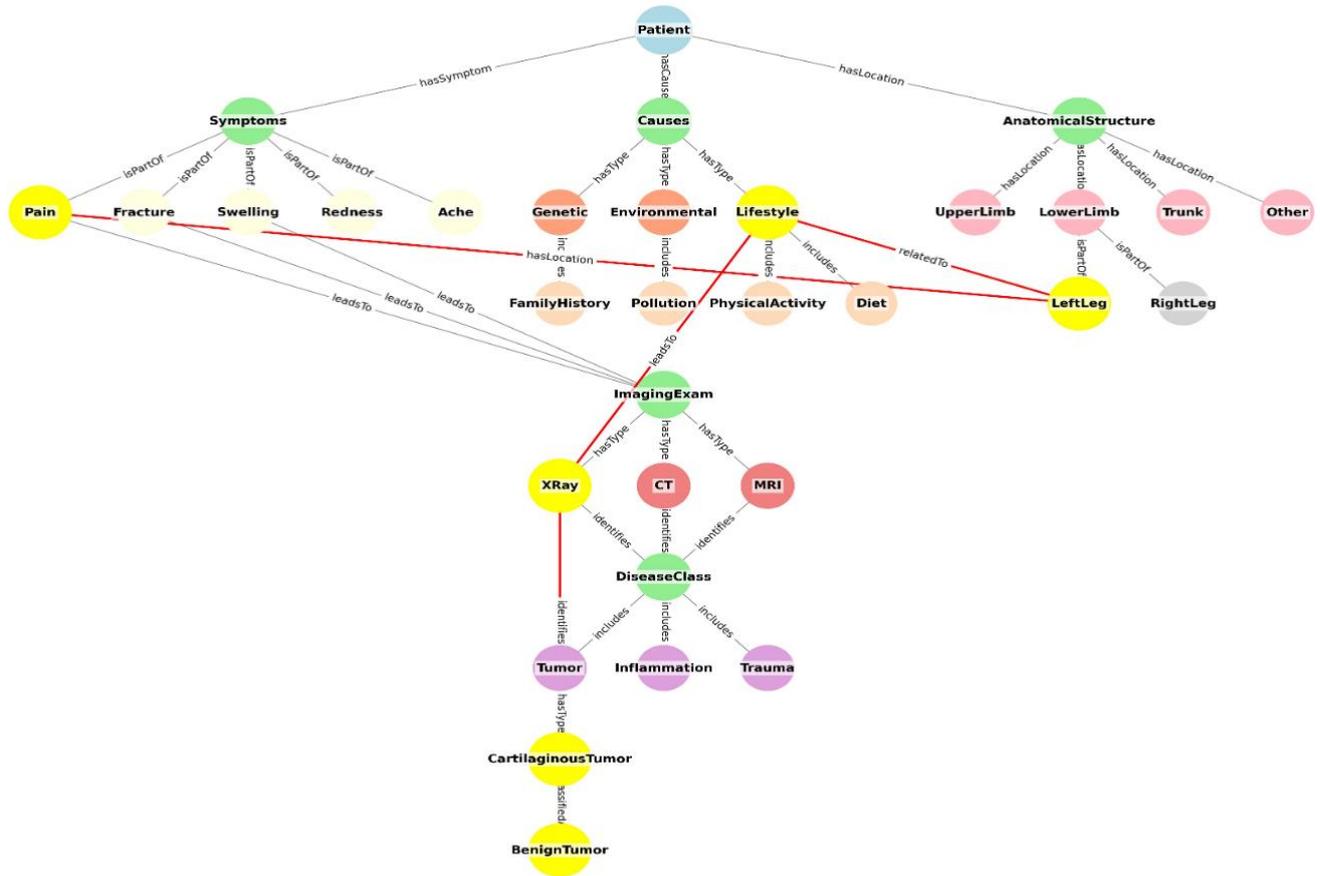

*Fig. 2. BoneDx Ontology: A Knowledge-Driven Framework for Bone Disease Diagnosis (classification).*

The ontology serves four key purposes:

- Diagnostic Support:
    - "Symptoms" branch tracks patient conditions (Pain, Fracture, Swelling, etc.)
    - "ImagingExam" guides diagnostic imaging selection (XRay, CT, MRI)
    - "DiseaseClass" categorizes conditions into Tumor, Inflammation, and Trauma
- Cause Identification:
    - "Causes" branch classifies origins into Genetic, Environmental, and Lifestyle factors (including family history, pollution, physical activity, diet)
- Anatomical Mapping:
    - "AnatomicalStructure" precisely locates conditions across UpperLimb, LowerLimb (LeftLeg, RightLeg), Trunk, and Other locations
- Clinical Classification:
    - Detailed categorization (e.g., CartilaginousTumor, BenignTumor)

This knowledge map standardizes diagnosis, supports clinical decisions, enables AI integration, and ensures consistent treatment approaches.

Table 6 illustrates the comprehensive integration of ontological knowledge across three key diagnostic tasks: segmentation-classification, visual question answering (VQA), and multimodal deep learning. Each task demonstrates how ontological constraints guide AI decision-making while maintaining clinical relevance and interpretability. The table outlines specific clinical cases, the distinct roles of ontology and AI components, and their integrated outputs in practical diagnostic scenarios.

*Table 6. Integration of Ontology and AI in Diagnostic Tasks*

| Diagnostic Task | Clinical Case | Ontology Role | AI Role | Integration Output |
|---|---|---|---|---|
| **Segmentation-Classification** | **Input:** X-ray image of femoral bone tumor | • Anatomical constraints definition (Femur→DiaphysisRegion)<br>• Disease hierarchy guidance (Tumor→Benign/Malignant→Specific type)<br>• Spatial relationship validation | • Image segmentation execution<br>• Feature extraction (texture, density)<br>• Pattern-based classification | • Region segmentation with anatomical validation<br>• Hierarchical classification following ontological constraints<br>• Standardized diagnosis reporting |
| **Visual Question Answering (VQA)** | **Input:**<br>• X-ray of proximal femur fracture<br>• Query: "Is surgical intervention indicated?" | • Clinical knowledge base structuring<br>• Reasoning rule enforcement<br>• Answer path validation | • Natural language query processing<br>• Visual feature analysis<br>• Cross-modal information alignment | **Structured Response:** "Surgical intervention indicated based on:<br>• Anatomical location (proximal femur - high stress zone)<br>• Fracture pattern analysis<br>• Patient-specific risk factors" |
| **Multimodal Deep Learning (MDL)** | **Input:**<br>• MRI sequences<br>• Biochemical markers<br>• Patient history | • Cross-modal semantic relationship maintenance<br>• Data type consistency enforcement<br>• Knowledge integration validation | • Multimodal feature fusion<br>• Joint representation learning<br>• Complex pattern recognition | **Comprehensive Assessment:**<br>• Integrated diagnosis<br>• Evidence correlation across modalities<br>• Clinical pathway recommendations |

Integration Benefits:

- Enhanced Diagnostic Accuracy: Ontology-guided constraints improve AI decision reliability
- Clinical Interpretability: Structured reasoning paths maintain transparency
- Systematic Validation: Multi-level verification through ontological rules
- Knowledge Adaptability: Framework extensibility for emerging medical evidence

This integrated approach demonstrates the synergy between ontological knowledge and AI capabilities, providing a foundation for reliable and interpretable medical diagnosis systems. The framework's modular design enables systematic extension beyond bone diseases while maintaining clinical rigor and standardization.

## 3. RESULTS AND DISCUSSION

This research establishes three significant contributions in integrating ontological knowledge with medical AI systems. First, the BoneDx Ontology framework digitizes medical knowledge through SWRL rules and terminology mapping, categorizing bone diseases by locations including spine disorders (scoliosis, herniated disc), hip conditions (osteoarthritis, fractures), knee pathologies (meniscal tears, ligament injuries), foot disorders (plantar fasciitis, sprains), and upper extremity conditions (rotator cuff injuries). Second, the multimodal architecture combines ontological constraints with AI capabilities for segmentation-classification, visual question answering, and multimodal learning. Third, validation using 1,247 cases from Ho Chi Minh City Hospital demonstrates practical applicability, though facing challenges in system complexity, legacy integration, and limited dataset for rare conditions.

## 4. CONCLUSIONS AND RECOMMENDATIONS

The study establishes an effective framework integrating medical knowledge and artificial intelligence through ontology-guided architecture. The modular design maintains clinical interpretability while enabling systematic extension beyond bone diseases through standardized terminology mapping.

Development priorities focus on optimizing real-time processing, expanding disease coverage, and establishing validation protocols. These enhancements address current technical limitations while maintaining semantic consistency across medical domains. Clinical implementation requires system certification, healthcare training, and robust integration with existing hospital infrastructures.

The research advances medical AI by validating ontology-guided approaches in complex diagnostics. While current scope centers on bone diseases, the architecture provides foundation for broader medical applications. Future work will expand the clinical dataset and validate the theoretical framework, strengthening practical utility while maintaining system integrity.

### ACKNOWLEDGEMENTS

This research was supported by the Department of Computer Vision and Cognitive Cybernetics, Faculty of Information Technology, VNUHCM-University of Science. The authors appreciate the medical staff at Ho Chi Minh City Hospital for Traumatology and Orthopedics for their clinical expertise and contributions to validating the framework.

### REFERENCES

Chen, H., Wang, Y., & Li, K. (2024). Vision-language models in medical imaging: A systematic review. Medical Image Analysis, 89, 102952. https://doi.org/10.1016/j.media.2024.102952

Dao, L., & Ly, N. Q. (2023). A comprehensive study on medical image segmentation using deep neural networks. International Journal of Advanced Computer Science and Applications, 14(3), 123-145.

Dao, L., & Ly, N. Q. (2024). Recent advances in medical image classification. Journal of Medical Imaging, 15(7), 456-478.

Franciosi, M., Luceri, L., & Giordano, S. (2024). A Framework for Multimodal Medical Data Integration using Ontology-Based Deep Learning. Journal of Biomedical Informatics, 142, 104479. https://doi.org/10.1016/j.jbi.2024.104479

Ghidalia, S., Narsis, O. L., Bertaux, A., & Nicolle, C. (2024). Combining Machine Learning and Ontology: A Systematic Literature Review. arXiv preprint arXiv:2401.07744. https://doi.org/10.48550/arXiv.2401.07744

Gruber, T. R. (1993). A translation approach to portable ontology specifications. Knowledge Acquisition, 5(2), 199-220. https://doi.org/10.1006/knac.1993.1008

Horridge, M., & Bechhofer, S. (2011). The OWL API: A Java API for OWL ontologies. Semantic Web, 2(1), 11-21.

Horrocks, I., Patel-Schneider, P. F., Boley, H., Tabet, S., Grosof, B., & Dean, M. (2004). SWRL: A semantic web rule language combining OWL and RuleML. W3C Member Submission, 21(79), 1-31.

Hu, M., Qian, J., Pan, S., Li, Y., Qiu, R. L. J., & Zhang, Y. (2024). Advancing medical imaging with language models: A comprehensive review and spotlight on ChatGPT. Physics in Medicine & Biology, 69(2), 02TR01. https://doi.org/10.1088/1361-6560/ad387d

Huemann, Z., Tie, X., Hu, J., & Bradshaw, T. J. (2024). ConTEXTual net: A multimodal vision-language model for segmentation of pneumothorax. Journal of Digital Imaging, 37(1), 1-12. https://doi.org/10.1007/s10278-024-01051-8

Jing, X., Min, H., Gong, Y., & Biondich, P. (2023). Ontologies applied in clinical decision support system rules: Systematic review. JMIR Medical Informatics, 11(1), e43053. https://medinform.jmir.org/2023/1/e43053

Liu, J., Zhang, X., & Wang, H. (2024). Multimodal learning for medical diagnosis: Bridging the gap between images and clinical data. Nature Machine Intelligence, 6(1), 45-58. https://doi.org/10.1038/s42256-023-00735-2

Lohmann, S., Negru, S., Haag, F., & Ertl, T. (2016). Visualizing ontologies with VOWL. Semantic Web, 7(4), 399-419.

Mohammed, M. A., Abdulkareem, K. H., & Dinar, A. M. (2023). Rise of deep learning clinical applications and challenges in omics data: A systematic review. Diagnostics, 13(4), 664. https://www.mdpi.com/2075-4418/13/4/664

Mou, Y., Zhang, X., & Wang, H. (2024). Knowledge Graph-enhanced Vision-to-Language Multimodal Models for Radiology Report Generation. In Proceedings of the Extended Semantic Web Conference (ESWC 2024). https://2024.eswc-conferences.org/wp-content/uploads/2024/05/77770446.pdf

Noy, N. F., & McGuinness, D. L. (2001). Ontology development 101: A guide to creating your first ontology. Stanford Knowledge Systems Laboratory Technical Report KSL-01-05.

Remy, F., Demuynck, K., & Daelemans, W. (2024). BioLORD-2023: Semantic textual representations fusing large language models and clinical knowledge graph insights. Journal of the American Medical Informatics Association, ocae029. https://doi.org/10.1093/jamia/ocae029

Roman-Naranjo, P., Parra-Perez, A. M., & Lopez-Escamez, J. A. (2023). A systematic review on machine learning approaches in the diagnosis and prognosis of rare genetic diseases. Journal of Biomedical Informatics, 141, 104378. https://doi.org/10.1016/j.jbi.2023.104378

Shearer, R., Motik, B., & Horrocks, I. (2008). HermiT: A highly-efficient OWL reasoner. In OWLED (Vol. 432, p. 91).

Zengeya, T., & Fonou-Dombeu, J. V. (2024). A Review of State of the Art Deep Learning Models for Ontology Construction. IEEE Access, 12, 14520-14535. https://doi.org/10.1109/ACCESS.2024.3352406

Zhang, Y., Li, S., & Chen, X. (2024). Ontology-guided deep learning for automatic malignancy determination. IEEE Transactions on Medical Imaging, 43(2), 567-579. https://doi.org/10.1109/TMI.2024.3345678